%% file: emnlp2020.tex
\documentclass[11pt,a4paper]{article}
\usepackage[hyperref]{emnlp2020}
\usepackage{times}
\usepackage{latexsym}

\usepackage{microtype}
\usepackage{paralist}
\usepackage{amsmath}
\usepackage{mathrsfs,amssymb}
\usepackage{enumitem}
\usepackage{color}
\usepackage{graphicx}
\graphicspath{figures}
\usepackage{caption} 
\usepackage[labelformat=simple]{subcaption}

\usepackage{xspace}
\usepackage{array}
\usepackage{xurl}
\usepackage{multirow}
\usepackage{todonotes}

\aclfinalcopy % Uncomment this line for the final submission
\def\aclpaperid{2886} %  Enter the acl Paper ID here

\newcommand{\lb}{{local group bias}\xspace}
\newcommand{\gb}{{global group bias}\xspace}
\newcommand{\mtd}{\textsc{LOGAN}\xspace}

\title{LOGAN: Local Group Bias Detection by Clustering}

\author{Jieyu Zhao \quad Kai-Wei Chang
\\
  University of California, Los Angeles  \\
  jyzhao@cs.ucla.edu  \quad  kwchang@cs.ucla.edu
}

\date{}

\begin{document}
\maketitle
\begin{abstract}
\input{sections/abstract.tex}
\end{abstract}

\section{Introduction}
\input{sections/intro}

\section{Related Work}
\input{sections/related_work}

\section{Methodology}
\input{sections/method}

\section{Experiments}
\input{sections/exp}

\section{Conclusion}
\input{sections/dis}

\bibliographystyle{acl_natbib}
\bibliography{ucla,cited}

\clearpage
\appendix

\section{Appendices}
\label{sec:appendix}
\input{sections/appendix}

\end{document}

%% file: sections/abstract.tex
Machine learning techniques have been widely used in natural language processing (NLP). However, as revealed by many recent studies, machine learning models often inherit and amplify the societal biases in data. 
Various metrics have been proposed to quantify biases in model predictions.
In particular, several of them evaluate disparity in model performance between protected groups and advantaged groups in the test corpus. 
However, we argue that evaluating bias at the corpus level is not enough for understanding how biases are embedded in a model. 
In fact, a model with similar aggregated performance between different groups on the entire data may behave differently on instances in a local region.  To analyze and detect such  \emph{local bias}, we propose \mtd, a new bias detection technique based on clustering. Experiments on toxicity classification and object classification tasks show that \mtd identifies bias in a local region and allows us to better analyze the biases in model predictions. 

%% file: sections/intro.tex
Machine learning models such as deep neural networks have achieved remarkable performance in many NLP tasks. However, as noticed by recent studies, these models often inherit and amplify the biases in the datasets used to train the models
\cite{zhao2017men, bolukbasi2016man, caliskan2017semantics,zhou2019grammaticalgenderbias,manzini2019black,blodgett-etal-2020-language}.  

To quantify bias, researchers have proposed various metrics to study algorithmic fairness at both individual and group levels. The former measures if a model treats similar individuals consistently no matter which groups they belong to, while the latter requires the model to perform similarly for protected groups and advantaged groups in the corpus.\footnote{For example, \citet{zhao2018gender} and \citet{rachel18} evaluate the bias in coreference resolution systems by measuring the difference in $F_1$ score between cases where a gender pronoun refers to an occupation stereotypical to the gender and the opposite situation.}
 In this paper, we argue that studying algorithmic fairness at either level does not tell the full story.  A model that reports similar performance across two groups in a corpus may behave differently between these two groups in a local region. 
 
For example, the performance gap of a toxicity classifier for sentences mentioning black and white race groups is $4.8\%$.\footnote{ 
 Performance in accuracy on the unintended bias detection task \cite{jigsaw}} This gap is only marginally larger than the performance gap of 2.4\% when evaluating the model on two randomly split groups. 
However, if we evaluate the performance gap on the sentences containing the token ``racist'', the performance gap between these two groups is as large as $19\%$. Similarly, \citet{zhao2017men} report that a visual semantic role labeling system tends to label an image depicting cooking as \emph{woman cooking} than \emph{man cooking}. However, the model is, in fact, more likely to produce an output of \emph{man cooking} when the agent in the image wears a chef hat. We call these biases exhibited in a neighborhood of instances \textbf{local group bias} in contrast with  \textbf{global group bias} which is evaluated on the entire corpus.

To detect \emph{local group bias}, we propose \mtd, a LOcal Group biAs detectioN algorithm to identify biases in local regions.
\mtd adapts a clustering algorithm (e.g., K-Means) to group instances based on their features while  maximizing a bias metric (e.g., performance gap across groups) within each cluster. In this way, local group bias is highlighted, allowing a developer to further examine the issue.

Our experiments on toxicity classification and MS-COCO object classification demonstrate the effectiveness of \mtd. We show that besides successfully detecting \lb, our method also provides interpretations for the detected bias. For example, we find that different topics lead to different levels of \lb in the toxicity classification.

%% file: sections/related_work.tex
\paragraph{Bias Evaluation}
Researchers have proposed to study algorithmic fairness from both individual and group perspectives~\cite{dwork2012fairness,dwork2018group}. 
To analyze group fairness, various metrics have been proposed. For example, demographic parity~\cite{dwork2012fairness} requires the probability of the predictor making positive prediction to be independent of the sensitive attributes. However this metric cannot always guarantee fairness, as we can  accept  correct examples in one demographic group but  make random guess in another one as long as we maintain the same acceptance ratio. To solve this problem, \citet{hardt2016equality} propose new metrics, equalized odds and equalized opportunity, to measure the discrimination related to the sensitive attributes which require the predictions to be independent of the demographic attributes given true labels. In NLP, many studies use the performance gap between different demographic groups as a bias measurement~\cite{gaut2019towards,kiritchenko2018examining,wang2019iccv}. The choice of bias metric depends on applications. In this work, we use performance gap as the bias evaluation metric. However, our approach can be generalized to other metrics.

\paragraph{Bias in NLP Applications} Recent advances in machine learning models boost the performance of various NLP applications. However, recent studies show that biases exhibit in NLP models. For example, researchers demonstrate that representations in NLP models are biased toward certain societal groups \cite{bolukbasi2016man, caliskan2017semantics, zhao2018learning,zhao2019gender,zhou2019grammaticalgenderbias, may2019measuring}. 
\citet{stanovsky2019evaluating} and \newcite{font2019equalizing} show that gender bias exhibits in neural machine translations while \citet{dixon2018measuring} and \citet{sap2019risk} reveal biases in text classification tasks. Other applications such as cross-lingual transfer learning \cite{zhao2020gender} and natural language generation \cite{sheng2019woman} also exhibit unintended biases.

%% file: sections/method.tex
In this section, we first provide formal definitions of \lb and then the details of the detection method \mtd.
\paragraph{Performance Disparity}
 Assume we have a trained model $f$ and a test corpus $\mathcal{D}=\{(x_i, y_i)\}_{i=1\dots n}$ that is used to evaluate the model. Let $P_f(\mathcal{D})$ represents the performance of the model $f$ evaluated on the corpus $\mathcal{D}$. 
Based on the applications, the performance metric can be accuracy, AUC, false positive rates, etc. 
For the sake of simplicity, we assume each input example $x_i$ is associated with one of demographic groups (e.g., male or female), i.e.,  $x_i \in \mathcal{A}_1$ or $x_i \in \mathcal{A}_2$.\footnote{In this paper, we  consider only binary attributes such as gender = \{male, female\}, race = \{white, black\}. However, our approach is general and can be incorporated with any bias metric  presented as a loss function. Therefore, it can be straightforwardly extended to a multi-class case by plugging the corresponding bias metric.
} As a running example, we take performance disparity as the bias metric. That is, if $\|P_f(\mathcal{A}_1) - P_f(\mathcal{A}_2)\| > \epsilon$, then we consider that the model exhibits bias, where $\epsilon$ is a given threshold.

\paragraph{Definition of local group bias}
We define \emph{local group bias} as the bias exhibits in certain local region of the test examples. Formally, given a centroid $c$ in the input space, let $\mathcal{A}^c_1 = \{x \in \mathcal{A}_1 | \|x - c \|^2  < \gamma \}$ and $\mathcal{A}^c_2 = \{x \in \mathcal{A}_2 | \|x - c \|^2  < \gamma \}$ be the neighbor instances of $c$ in each group, where $\gamma$ is a threshold. 
We call a model has \lb if 
\begin{equation}
\label{eq:perfgap}
\|P_f(\mathcal{A}^c_1) - P_f(\mathcal{A}^c_2)\| > \epsilon.
\end{equation}
While this definition is based on performance disparity, it is straightforward to extend the notion of \lb to other bias metrics.

\paragraph{\mtd }
The goal of \mtd is to cluster instances in $\mathcal{D}$ such that (1) similar examples are grouped together, and (2) each cluster demonstrates local group bias contained in $f$. To achieve this goal, \mtd generates cluster $\mathcal{C}=\{C_{i,j}\}_{i=1\ldots n, j=1\ldots k}$ by optimizing the following objective:
\begin{equation}
    \label{eq:final}
    \min\nolimits_{\mathcal{C}} L_c + \lambda L_b,
\end{equation}
where $L_c$ is the clustering loss and $L_b$ is local group bias loss.
$\lambda \ge 0$ is a hyper-parameter to control the trade-offs between the two objectives. $C_{ij}=1$ if $x_i$ is assigned to the cluster $j$; $C_{ij}=0$ otherwise.
We introduce these two loss terms in the following. 

\paragraph{Clustering objective} The loss $L_c$ is derived from a standard clustering technique. In this paper, we consider the 
K-Means clustering method~\cite{lloyd1982least}. Specifically, the loss $L_c$ of K-Means is 
\begin{equation}
\label{eq:kmeans}
    L_c = \sum_{j=1}^{k}  \sum_{i=1}^{n} \|C_{ij}x_i - \mu_j\|^2  \quad
     \forall i, \sum_{j=1}^{k} C_{ij} = 1, 
\end{equation}
 $\mu_j=(\sum_{ij}C_{ij}x_i)/\sum_{i,j}C_{ij}$ is the mean of cluster $j$. Note that our framework is general and other clustering techniques, such as Spectral clustering~\cite{shi2000normalized}, DBSCAN~\cite{ester1996density}, or Gaussian mixture model can  also be applied in generating the clusters. Besides, the features used for creating the clusters can be different from the features used in the  model $f$.

\paragraph{Local group bias objective}
For the local group bias loss $L_b$, the goal is to obtain a clustering that maximizes the bias metric within each cluster. In the following descriptions, we take the performance gap between different attributes (see Eq. \eqref{eq:perfgap}) as an example to describe the bias metric. 

Let $\hat{y}_i=f(x_i)$  be the prediction of $f$ on $x_i$. The local group bias loss $L_b$ is defined as the negative summation of performance gaps over all the clusters. If accuracy is used as the performance evaluation metric, $L_b=$ 
\begin{equation*}
    \label{eq:lb}
%    \begin{split}
        - \sum_{j=1}^{k}  \left| \frac{\sum_{x_i\in \mathcal{A}_1}C_{ij}\mathcal{I}_{\hat{y}_i = y_i}}{\sum_{x_i\in \mathcal{A}_1} C_{ij}} -  \frac{\sum_{x_i\in \mathcal{A}_2} C_{ij}\mathcal{I}_{\hat{y}_i = y_i}}{\sum_{x_i\in \mathcal{A}_2} C_{ij}}  \right|^2,
%    \end{split}
\end{equation*}
where $\mathcal{I}$ is the indicator function.

Similar to K-Means algorithm,  we solve Eq.~\eqref{eq:final} by iterating two steps: first, assign $x_i$ to its closest cluster $j$ based on current $\mu_j$; second, update $\mu_j$ based on current label assignment. We use k-means++~\cite{arthur2007k} for the cluster initialization and stop when the model converges or reaches enough iterations. To make sure each cluster contains enough instances, in practice, we choose a large $k$ ($k=10$ in our case) and merge a small cluster to its closest neighbor.~\footnote{We merge the clusters iteratively and stop the procedure when all the clusters have at least 20 examples or  only 5 clusters are left.} For \lb detection, we only consider  clusters with at least 20 examples from each group.

%% file: sections/exp.tex
In this section, we show that \mtd is capable of identifying local group  bias, and the clusters generated by \mtd provide an insight into how bias is embedded in the model. 

\subsection{Toxicity Classification}
This task aims at detecting whether a comment is toxic (e.g. abusive or rude). Previous work has demonstrated that this task is biased towards specific identities such as ``gay'' \cite{dixon2018measuring}. In our work, we use toxicity classification as one example to detect \lb in texts and show that such local group bias could be caused by different topics in the texts.   
\paragraph{Dataset} We use the official train and test datasets from \citet{jigsaw}. As the dataset is extremely imbalanced, we down-sample the training dataset and reserve 20\% of it as the development set. In the end, we have $204,000$, $51,000$ and $97,320$ examples for train, development and test, respectively.  We tune  $\lambda = \{1, 5, 10, 100\}$ and choose the one with the largest number of clusters showing local group bias.

\paragraph{Model} We fine-tune a BERT sequence classification model from~\citet{Wolf2019HuggingFacesTS}  for 2 epochs with a learning rate $2\times10^{-5}$, max sequence length $220$ and batch size $20$.  The model achieves $90.2\%$ accuracy on the whole test dataset.\footnote{The source code is available at \url{https://github.com/uclanlp/clusters}.} 
We use sentence embeddings from the second to last layer of a pre-trained BERT model as features to perform clustering. We also provide clustering results based on the sentence embeddings extracted from a fine-tuned model in Appendix~\ref{app:res_finetuned}.

\begin{figure}[!t]
\centering
    \includegraphics[width = 0.8\linewidth]{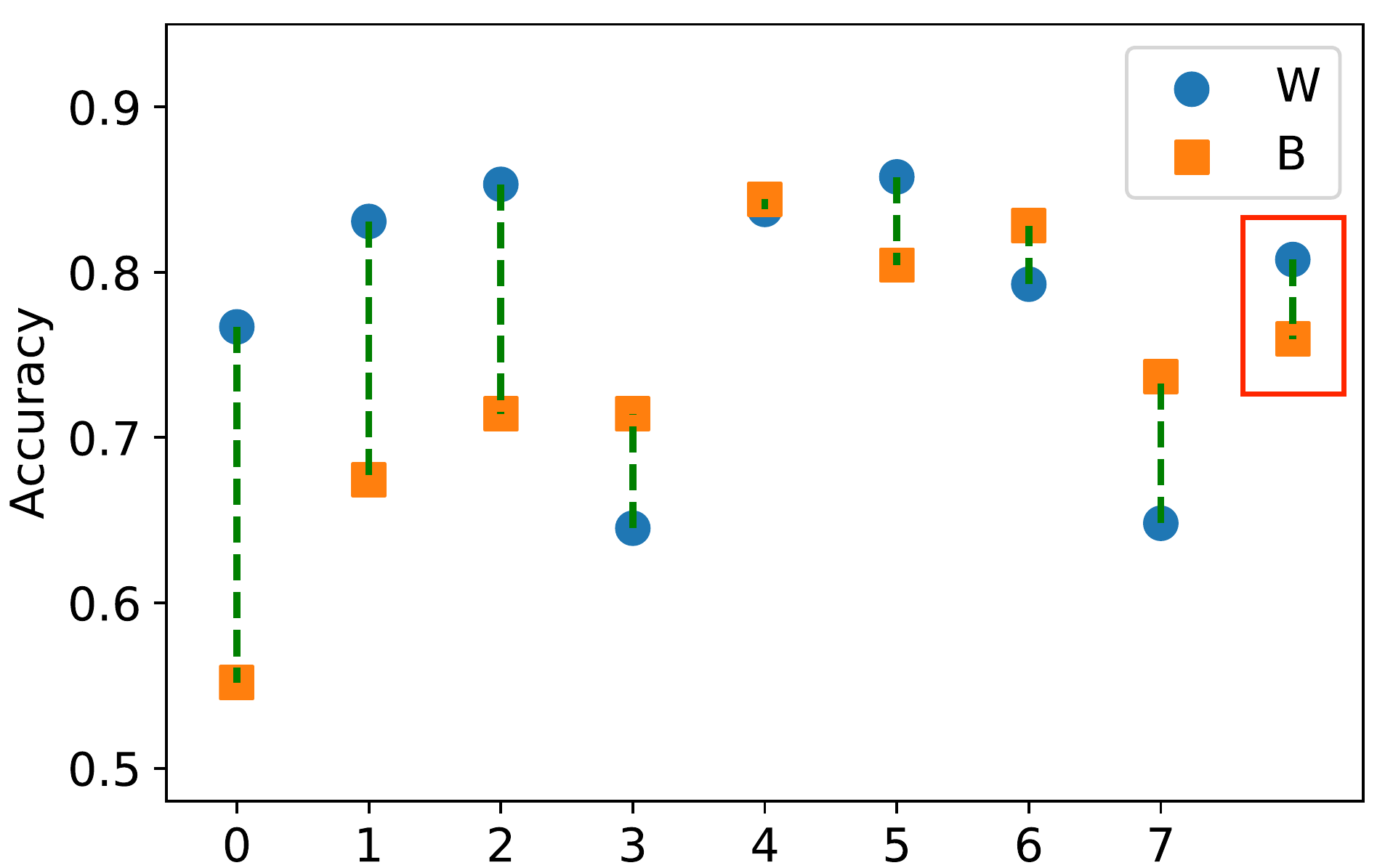}
\caption{Accuracy for White (blue circle) and Black (orange square) groups in each cluster using \mtd. The length of the dashed line shows the gap. Red box highlights the accuracy of these two groups on the entire corpus. Clusters 0 and 1 demonstrate strong local group bias. 
Full results are in Appendix~\ref{app2}.} 
\label{fig:localbias}
\end{figure}

\begin{table}[!t]
\centering
\begin{tabular}{@{}l@{}|c|c|c|c}
\hline
 \multirow{4}{*}{\textsc{Race}}  & Method & Acc-W & Acc-B & $\mid$Bias$\mid$\\
\cline{2-5}
& Global   & 80.8 & 76.0 & 4.8  \\
& K-Means &  75.9 & 53.8  & \textbf{22.1}  \\
&\mtd  & 76.7  & 55.2 & \textbf{21.5} \\
\hline
\hline
\multirow{4}{*}{\textsc{Gender}}  & Method & Acc-M & Acc-F & $\mid$Bias$\mid$ \\
 \cline{2-5}
& Global      &  79.8    &  81.6  & 1.8  \\
& K-Means &  70.2  &  82.8 &  \textbf{12.6} \\
& \mtd & 80.2  &  57.1 & \textbf{23.1}  \\
\hline
\end{tabular}
\caption{Bias detection in toxic classification. Results are shown in \%. ``Global'' stands for  \gb detection. W, B, M, F refer to White, Black, Male and Female groups respectively.}
\label{tab:bw}
\end{table}

\paragraph{Bias Detection} 
There are several demographic groups in the toxic dataset such as gender, race and religion. We focus on the binary gender (male/female) and binary race (black/white) in the experiments. 
 For \lb, we report the largest bias score among all the clusters. Figure~\ref{fig:localbias} shows the accuracy  of white and black groups in each cluster using \mtd. The example bounded in the red box is the global accuracy of these two groups.
 Based on the results in  Figure~\ref{fig:localbias} and Table \ref{tab:bw}, we only detect weak \gb in the model predictions. However, 
both K-Means and \mtd  successfully detect strong \lb. In particular, \mtd identifies a local region that the model has difficulties in making correct predictions for female group.

While we use the gap of accuracy as the bias metric, the clusters detected by \mtd also exhibit local bias when evaluating using other metrics. Table~\ref{tab:bw_auc} shows the gap of subgroup AUC scores over the clusters. Similar to the results in Table \ref{tab:bw}, K-Means and \mtd  detect \lb. In particular, the first and the third clusters in Figure~\ref{fig:localbias} also have larger AUC disparity than the global AUC gap.
Similarly, the first three clusters in Figure~\ref{fig:localbias} have a significantly larger gap of False Positive Rate across different groups than when evaluating on the entire dataset.

\begin{table}[!t]
\centering
\begin{tabular}{@{}l@{}|c|c|c|c}
\hline
 \multirow{4}{*}{\textsc{Race}}  & Method & AUC-W & AUC-B & $\mid$Bias$\mid$\\
\cline{2-5}
& Global   & 0.870 & 0.846 & 0.024  \\
& K-Means &  0.836 & 0.679  & \textbf{0.157}  \\
&\mtd  & 0.844 & 0.691 & \textbf{0.153} \\
\hline
\hline
\multirow{4}{*}{\textsc{Gender}}  & Method & AUC-M & AUC-F & $\mid$Bias$\mid$ \\
 \cline{2-5}
& Global      &  0.896    &   0.924  &0.028\\
& K-Means &   0.828  &  0.922 &  \textbf{0.094} \\
& \mtd &  0.910  &  0.818  & \textbf{0.092 }  \\
\hline
\end{tabular}
\caption{Bias detection using subgroup AUC. ``Global'' stands for  \gb detection. W, B, M, F refer to White, Black, Male and Female groups respectively.}
\label{tab:bw_auc}
\end{table}

\paragraph{Bias Interpretation}
To  better interpret the \lb, we run a Latent Dirichlet Allocation topic model~\cite{blei2003latent} to discover the main topic of each cluster. 
Table~\ref{tab:itopics_race} lists the top 20 topic words for the most and least biased clusters using \mtd under \textsc{race} attributes. We remove the words related to  race attributes such as ``white'' and ``black''.
Other results are in Appendix~\ref{app:topics}. We find that different topics in each cluster may lead to different levels of \lb. For example, compared with the less biased group, the most biased group includes a topic on supremacy.

\begin{table}[!t]
\centering
\begin{tabular}{l|l}
\hline
\multirow{5}{*}{\shortstack[c]{Most \\ Biased \\ (21.5)}} &  trump  supremacist supremacists kkk \\
& people party america racist \\
& president support vote sessions \\
& voters republican said obama \\
& man base bannon nationalists  \\ 
\hline
\multirow{4}{*}{\shortstack[c]{Least \\ Biased \\ (0.6)}} &    people   like get think  know\\
&  say men see racist way\\
&  good point right go person\\
&  well  make time said much\\
\hline
\end{tabular}
\caption{Top 20 topic words in the most and least biased cluster using \mtd under \textsc{RACE} attributes. Number in parentheses is the bias score (\%) of that cluster.}
\label{tab:itopics_race}
\end{table}

\paragraph{Comparison between K-Means and \mtd}
We compare \mtd with K-Means using the following 3 metrics. ``Inertia'' sums over the distances of all instances to their closest centers which is used to measure the clustering quality. We normalize it to make the inertia of K-Means  $1.0$. To measure the utility of \lb detection, we look at the ratio of clusters showing a bias score at least $5\%$\footnote{We choose $5\%$ as it is close to the averaged bias score plus standard deviation when we randomly split the examples into two groups over 5 runs.} (BCR) as well as the ratio of instances within those biased clusters (BIR). 
Table~\ref{tab:compare} shows that \mtd increases the ratio of clusters exhibiting non-trivial \lb by a large margin with trivial trade offs in inertia.

\begin{table}[!t]
\centering
\begin{tabular}{c|c|c|c|c}
\hline
  & Inertia    & BCR & BIR  & $\mid$Bias$\mid$ \\
\hline
K-Means    & 1.0   &  62.5\%  & 58.2\% &   12.4\% \\
\mtd   &  1.002  &  75.0\%  & 71.8\% &   12.0\%\\
\hline
\end{tabular}
\caption{Comparison between K-Means and \mtd under \textsc{race} attributes. `` BCR'' and ``BIR'' refer to the ratio of biased clusters and ratio of instances in those biased clusters, respectively. ``$\mid$Bias$\mid$'' here is the averaged absolute bias score for those biased clusters.}
\label{tab:compare}
\end{table}

\subsection{Object Classification}
We conduct experiments on  object classification using MS-COCO \cite{lin2014microsoft}. Given one image, the goal is to predict if one object appears in the image. Following the setup in \citet{wang2019iccv}, we exclude person from the object labels.
\paragraph{Dataset} Similar to \citet{zhao2017men} and \citet{wang2019iccv}, we extract the gender label for one image by looking at the captions. For our analysis, we only consider images with gender labels.  In the end, there are $22,800$, $5,400$ and $5,400$ images left for train, development and test, respectively. 

\paragraph{Model} We use the basic model from \citet{wang2019iccv} for this task, which adapts a standard ResNet-50 pre-trained  on ImageNet  with the last layer modified.  We follow the default hyper-parameters of the original model.

\paragraph{Bias Detection and Interpretation}
We  evaluate bias in the predictions of the object classification model by looking at the accuracy gap between male and female groups for each object. 
In the analysis, we only consider objects with more than 100 images in the test set. 
This results in a total of 26 objects. Among the three methods, Global can only detect group bias at threshold $5\%$ (i.e., performance gap $\ge 5\%$) for 14 objects, while K-Means and \mtd increase the number to 19 and 21 respectively. 

Comparing \mtd with K-Means, among all the 26 objects, the average inertia is almost the same (the ratio is 1.001). On average, 34.0\% and 35.7\% of the clusters showing \lb at threshold $5\%$ (i.e. BCR) and the ratio of instances in those biased clusters (i.e., BIR) are 57.7\% and 54.9\% for K-Means and \mtd, respectively. 

We further investigate the local groups discovered by \mtd by comparing the images in the less biased local groups with the strong biased ones. We find that, for example, in the most biased local groups, the images often contain ``handbag'' with a street scene. In such a case, the model is more likely to correctly predict the agent in the image is woman (see Appendix~\ref{app:coco_localgroup}).

%% file: sections/dis.tex
Machine learning models risk inheriting the underlying societal biases from the data. In practice, many works use the global performance gap between different groups as a metric to detect the bias. 
In this work, we revisit the coarse-grained metric for group bias analysis and 
 propose a new method, \mtd, to detect local group bias by clustering. Our method can help  detect model biases that previously are hidden from the global bias metrics and provide an explanation of such biases. 
 
 We  notice there are some limitations in \mtd. For example, the number of instances in clusters could be uneven (see Appendix~\ref{app2}). 

\section*{Acknowledgment}
 This work was supported in part  by  National  Science  Foundation  Grant  IIS-1927554. We thank all the reviewers and members of UCLA-NLP and Plus labs for their feedback.

%% file: sections/appendix.tex
\subsection{Reproducibility}
We describe the details of our two models here. For toxicity classification tasks, we run the model on a GeForce GTX 1080 Ti GPU for 2 epochs, which takes about 3 hours to finish the fine-tuning procedure. The accuracy for the dev dataset is 89.4 \%.
For MS-COCO object classification tasks, we use the basic model from \url{https://github.com/uvavision/Balanced-Datasets-Are-Not-Enough}. We train the model based on the default hyperparameters in this repo (for example, batch size is $32$, learning rate is $10^{-5}$). We get meanAP 52.3\% and 53.1\% for development and test, respectively. We attach partial code in the supplemental materials. 

% =======

\subsection{Topic words in different clusters}
\label{app:topics}
We list all the top 20 words from the topic model using K-Means and \mtd in Table~\ref{tab:topics}, \ref{tab:topics_race} and \ref{tab:itopics_gender}.

\begin{table}[!h]
\centering
\small
\begin{tabular}{l|l}
\hline
\multirow{5}{*}{\shortstack{Most biased \\ (12.6) } }&  white black  people  like \\
& children  abortion right get  \\
&priests church take  canada  \\
& trump day think make\\
& young countries abortions time \\
\hline
\multirow{5}{*}{\shortstack{Least biased \\ (0.4)}} & https http  white trump   \\
&  like abortion muslim years  \\
& people religion  time know  \\
& read obama number go\\
& percent new said abortions \\
\hline
\end{tabular}
\caption{Top 20 words from the topic model for the most and least biased cluster using ``K-Means'' under \textsc{gender} attribute. Number in parenthese stands for the bias score of this cluster. }
\label{tab:topics}
\end{table}

\begin{table}[!h]
\centering
\small
\begin{tabular}{l|l}
\hline
\multirow{5}{*}{\shortstack[c]{Most \\ Biased \\ (22.1)}} &   trump supremacist supremacists  \\
& people kkk racist party sessions \\
& support america president vote \\
& said voters republican hate \\
& bannon right groups nazi \\
\hline
\multirow{4}{*}{\shortstack[c]{Least \\ Biased \\ (0.03) }} &    people  like get think go \\
&  know say men make person\\
&  right way good time well \\
&  see racist point said race\\
% &  hate \\
\hline
\end{tabular}
\caption{Top 20 topic words the most and least biased cluster using ``K-Means'' under \textsc{RACE} attributes. Number in parentheses is the bias score(\%) of that cluster.}
\label{tab:topics_race}
\end{table}

\begin{table}[!h]
\centering
\small
\begin{tabular}{l|l}
\hline
\multirow{5}{*}{\shortstack[c]{Most   Biased\\ (23.1)} } &  people like abortion think\\
&  know trump right get \\
& time make way  see\\
&  say said much care\\
&  well life go right\\ 
\hline
\multirow{5}{*}{\shortstack[c]{Least   Biased\\ (5.0)}} & people like trump get \\
& church right know think \\
& time never way see \\
& years make children go \\
& abortion  say rights good \\
\hline
\end{tabular}
\caption{Top 20 words from the topic model for the most and least biased cluster using \mtd under \textsc{gender} attributes. }
\label{tab:itopics_gender}
\end{table}

\iffalse
\begin{table}[!ht]
\centering
\begin{tabular}{l|l}
\hline
\multirow{5}{*}{\shortstack[c]{Most \\ Biased \\ (21.5)}} &  trump  supremacist supremacists kkk \\
& people party america racist \\
& president support vote sessions \\
& voters republican said obama \\
& man base bannon nationalists  \\ 
\hline
\multirow{5}{*}{\shortstack[c]{Least \\ Biased \\ (0.6)}} &    people   like get think \\
& know say men see racist \\
& way good point right \\
& go person well  make \\
& time said much going \\
\hline
\end{tabular}
\caption{Top 20 words from the topic model for the most and least biased cluster using ``iLocal'' under \textsc{race} attributes. }
\label{tab:itopics_race}
\end{table}
\fi

%=====================
\subsection{Local Bias Detection}
\label{app2}
Table~\ref{tab:c_Local} and \ref{tab:c_iLocal} list the results from the two clustering methods.

\begin{table}[!h]
\centering
\small
\begin{tabular}{l|c|c|c|c|c|c}
\hline
\multirow{10}{*}{\textsc{Gender}} &ID  & \#M  & \#F & M-acc & F-Acc & $\mid \text{Bias}\mid$\\
\cline{2-7}
 & 0  & 188 & 146 & 80.3 & 77.4 & 2.9 \\
 & 1 & 144 & 103 & 86.1 & 85.4 & 0.7 \\
 & 2 & 94 & 89 &  88.3 &  91.0 & 2.7 \\
 & 3 & 189 & 193 & 77.2 & 76.7 & 0.5 \\
 & 4 & 144 & 231 & 75.0 & 82.3 & 7.3 \\
 & 5 & 202 & 319 & 83.7 & 85.9 & 2.2 \\
 & 6 & 38 & 39 & 84.2 & 89.7 & 5.5 \\
 & 7 & 124 & 244 & 70.2 & 82.8 & 12.6 \\
 & 8 & 232 & 272 & 77.2 & 74.5 & 2.7 \\
 & 9 & 41 & 40 & 85.4 & 85.0 &  0.4 \\
\hline
\hline
\multirow{10}{*}{\textsc{Race}} &ID  & \#W  & \#B & W-acc & B-Acc & $\mid \text{Bias}\mid$\\
\cline{2-7}
 & 0 & 112 & 26 & 75.9 & 53.8 & 22.1 \\
 & 1 & 116 & 41 & 81.0 & 70.7 & 10.3 \\
 & 2 & 96 & 53 & 84.4 & 67.9 & 16.5 \\
 & 3 & 128 & 59 & 72.7 & 72.9 & 0.2 \\
 & 4 & 128 & 81 & 85.2 & 85.2  & 0\\
 & 5 & 192 & 75 & 88.0 & 80.0  & 8.0 \\
& 6 & 122 & 66 & 80.3 &  81.8 &  1.5 \\
& 7 & 63 & 40 & 69.8 & 75.0  & 5.2 \\ 
\hline
\end{tabular}
\caption{Bias detection on toxic classification using K-Means. Accuracy is shown in \%.}
\label{tab:c_Local}
\end{table}

\begin{table}[!h]
\centering
\small
\begin{tabular}{l|c|c|c|c|c|c}
\hline
\multirow{8}{*}{\textsc{Gender}} &ID  & \#M  & \#F & M-acc & F-Acc & $\mid \text{Bias}\mid$\\
\cline{2-7}
 & 0 & 245 & 29 & 82.9 & 75.9 & 7.0 \\
 & 1 & 172 & 41 & 78.5 & 63.4 & 15.1 \\
  & 2 & 176 & 626 & 80.1 & 85.1 & 5.0 \\
  & 3 & 212 & 70 & 78.3 & 64.3 & 14.0 \\
  & 4 & 294 & 787 & 79.6 & 85.0 & 5.4 \\
  & 5 & 216 & 52 & 78.7 & 61.5 & 17.2 \\
  & 6 &81 &  70 &  80.2 & 57.1 & 23.1 \\
\hline
\hline
\multirow{8}{*}{\textsc{Race}} &ID  & \#W  & \#B & W-acc & B-Acc & $\mid \text{Bias}\mid$\\
\cline{2-7}
 & 0 & 103& 29 & 76.7  & 55.2 & 21.5 \\
 & 1 & 130 & 43 & 83.1 & 67.4 & 15.6 \\
 & 2 & 109 & 56 & 85.3 & 71.4 & 13.9 \\
 & 3 &  62 & 42 & 64.5 & 71.4 &  6.9 \\
 & 4 &142 & 77 & 83.8 & 84.4 &  0.6 \\
 & 5 & 246 & 92 & 85.8 & 80.4 & 5.4 \\
 & 6 & 111 & 64 & 79.3 & 82.8 & 3.5 \\
 & 7 & 54 & 38 & 64.8 & 73.7 & 8.9 \\
\hline
\end{tabular}
\caption{Bias detection on toxic classification using \mtd. Accuracy is shown in \%.}
\label{tab:c_iLocal}
\end{table}

% =========================================================
\subsection{Results using embeddings extracted from a fine-tuned BERT model}
\label{app:res_finetuned}
In this section, we provide the results using the second to last layer embeddings from the fine-tuned BERT model to do local bias detection in Table~\ref{tab:clus_res_finetuned} and \ref{tab:iclus_res_finetuned}. 

\begin{table}[!h]
\centering
\small
\begin{tabular}{l|c|c|c|c|c|c}
\hline
\multirow{11}{*}{\textsc{Gender}} &ID  & \#M  & \#F & M-acc & F-Acc & $\mid\text{Bias}\mid$\\
\cline{2-7}
 & 0  & 88 & 52 & 98.9 & 100 & 1.1 \\
 & 1 & 155 & 140 & 95.5 & 98.6 & 3.1 \\
 & 2 & 60 & 46 &  88.3 &  87.0 & 1.3 \\
 & 3 & 237 & 362 & 99.2 & 99.2 & 0.0 \\
 & 4 & 184 & 255 & 96.2 & 95.7 & 0.5 \\
 & 5 & 130 & 191 & 26.2 & 31.9 & 5.8 \\
 & 6 & 101 & 129 & 66.3 & 67.4 & 1.1 \\
 & 7 & 169 & 192 & 99.4 & 99.5 & 0.1 \\
 & 8 & 114 & 44 & 46.5 & 43.2 & 3.3 \\
 & 9 & 158 & 264 & 58.2 & 66.7 &  8.4 \\
\hline
\hline
\multirow{8}{*}{\textsc{Race}} &ID  & \#W  & \#B & W-acc & B-Acc & $\mid\text{Bias}\mid$\\
\cline{2-7}
 & 0 & 221 & 75 & 91.5 & 89.3 & 2.2 \\
 & 1 & 81 & 47 & 60.5 & 59.6 & 0.9 \\
 & 2 & 253 & 103 & 97.2 & 97.1 & 0.1 \\
 & 3 & 165 & 71 & 78.8 & 76.1 & 2.7 \\
 & 4 & 96 & 50 & 59.4 & 48.0  & 11.4\\
 & 5 & 61 & 29 & 72.1 & 89.7  & 17.5 \\
& 6 & 90 & 66 & 60.0 & 54.5 &  5.4 \\
\hline
\end{tabular}
\caption{Local bias detection on toxic classification using K-Means. Accuracy is shown in \%.}
\label{tab:clus_res_finetuned}
\end{table}

\begin{table}[!h]
\centering
\small
\begin{tabular}{l|c|c|c|c|c|c}
\hline
\multirow{11}{*}{\textsc{Gender}} &ID  & \#M  & \#F & M-acc & F-Acc & $\mid\text{Bias}\mid$\\
\cline{2-7}
 & 0 & 31 & 342 & 45.2 & 78.1 &32.9 \\
 & 1 & 83 & 112 & 54.2 & 64.2 & 10.0 \\
 & 2 & 92 & 353 & 75.0 & 97.8 & 22.7 \\
 & 3 & 65 & 51 & 35.4 & 19.6 & 15.8 \\
 & 4 & 102 & 68 & 83.3 & 79.4 & 3.9 \\
 & 5 & 371 & 193 & 83.6 & 99.5 & 15.9 \\
 & 6 & 34 & 84 & 26.5 & 33.3 & 6.86 \\
 & 7 & 536 & 337 & 99.3 & 99.7 & 0.4 \\
 & 8 & 57 & 72 & 33.3 & 44.4 & 11.1 \\
 & 9 & 25 & 63 & 32.0 & 49.2 & 17.2\\
\hline
\hline
\multirow{10}{*}{\textsc{Race}} &ID  & \#W  & \#B & W-acc & B-Acc & $\mid\text{Bias}\mid$\\
\cline{2-7}
 & 0 & 24& 59 & 62.5  & 96.6 & 34.1 \\
 & 1 & 68 & 28 & 60.3 & 82.1 & 21.9 \\
 & 2 & 77 & 29 & 58.4 & 86.2 & 27.8 \\
 & 3 & 65 & 35 & 73.8 & 100 & 26.2 \\
 & 4 & 82 & 31 & 63.4 &90.3 & 26.9 \\
 & 5 & 466 & 92 & 90.8 & 95.7 & 4.9 \\
 & 6 & 35 & 63 & 85.7 & 49.2 & 36.5 \\
 & 7 &88 & 27 & 98.9 & 85.2 & 13.7 \\
 & 8 & 52 & 77 & 61.5 & 32.5 & 29.1\\
  
\hline
\end{tabular}
\caption{Local bias detection on toxic classification using \mtd. Accuracy is shown in \%.}
\label{tab:iclus_res_finetuned}
\end{table}

\subsection{Local Clusters for MS-COCO dataset}
\label{app:coco_localgroup}
In this section, we show the local group bias analysis for MS-COCO objection classification tasks.

\begin{figure*}[!t]
    \centering
    % \vspace{-10pt}
    % \hspace{-25pt}
    \begin{subfigure}[b]{\textwidth}
        \includegraphics[width=1\linewidth]{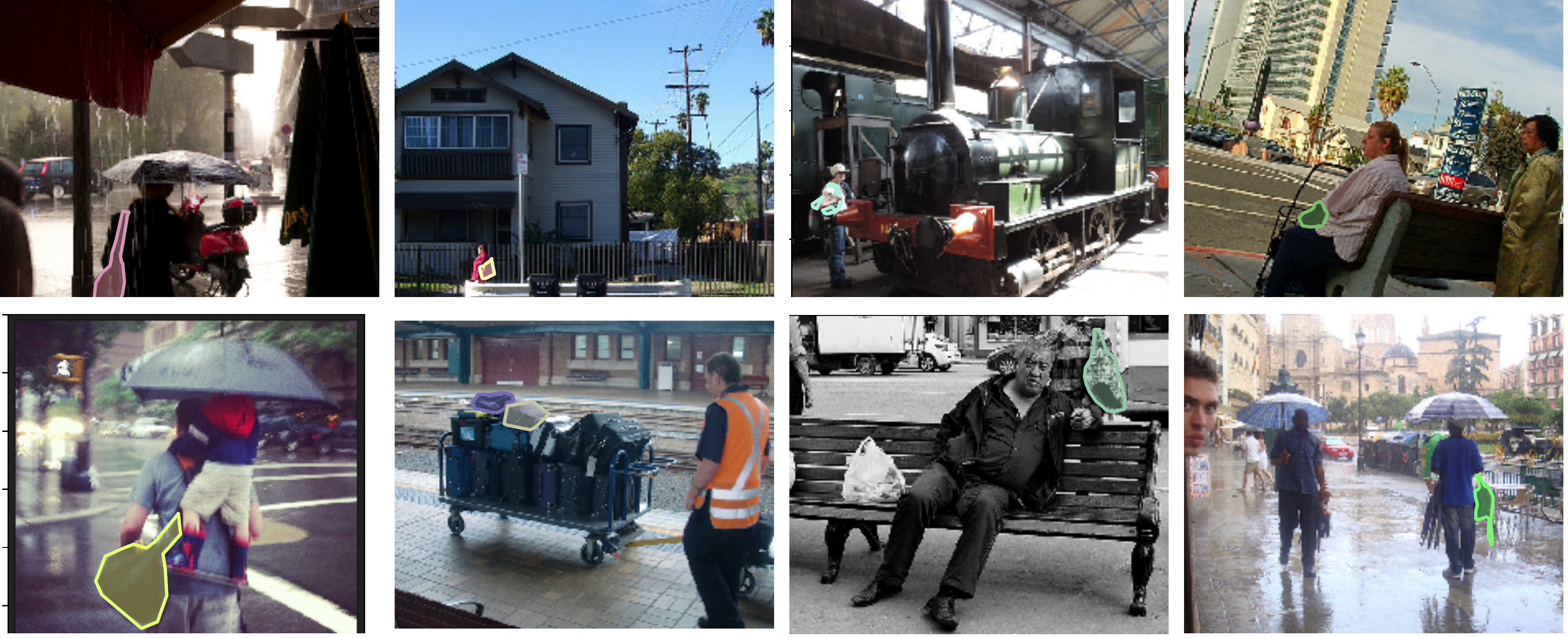}
    \end{subfigure}
    \begin{subfigure}[b]{\textwidth}
        \includegraphics[width=1\linewidth]{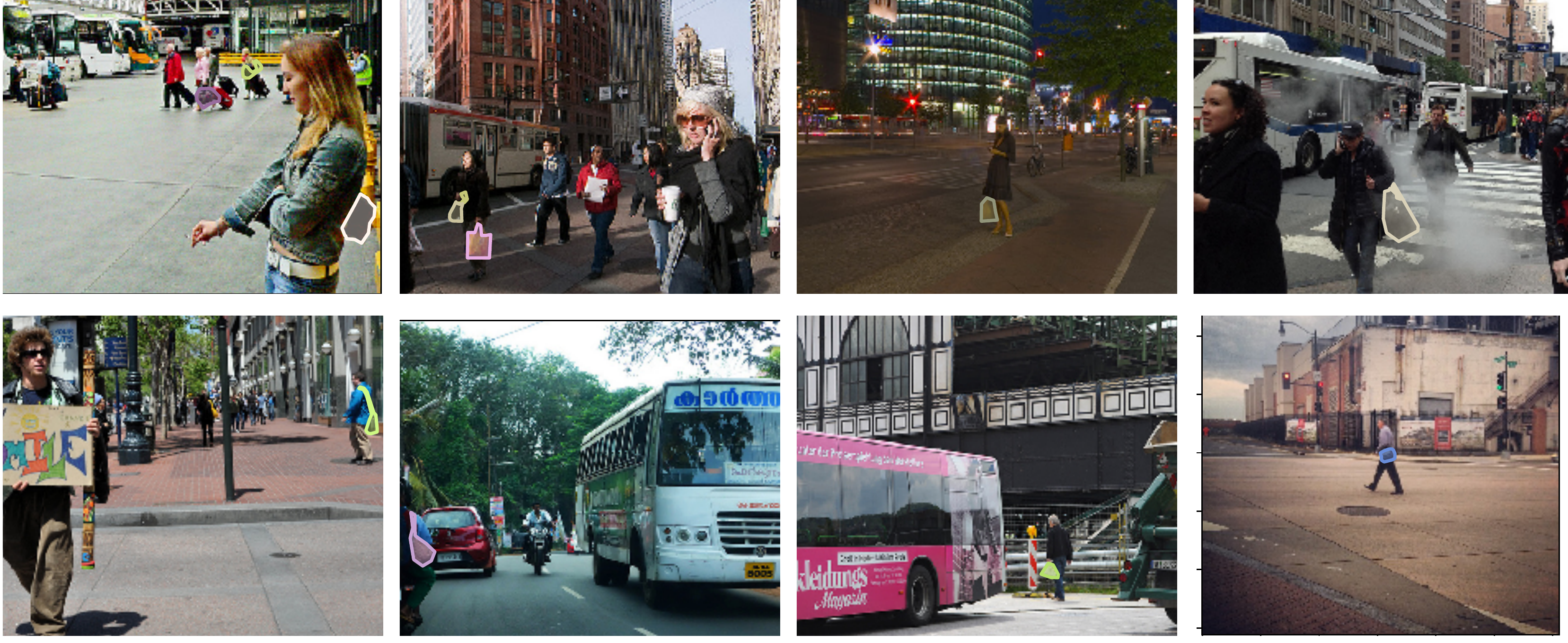}
    \end{subfigure}
    \caption{Images selected from least and most biased local groups using \mtd method. The top 2 and bottom 2 rows stand for the least and most biased clusters respectively. For each group, the first line is from female groups and the second line is from male groups.}
    \label{fig:coco_localgroup}
\end{figure*}